\pdfoutput=1

\documentclass[11pt]{article}

\usepackage[final]{acl}

\usepackage{latexsym}

\usepackage[T1]{fontenc}

\usepackage[utf8]{inputenc}

\usepackage{microtype}

\usepackage{inconsolata}

\usepackage{graphicx}

\usepackage{hyperref}       
\usepackage{url}            
\usepackage{booktabs}       
\usepackage{amsfonts}       
\usepackage{nicefrac}       
\usepackage{microtype}      
\usepackage{xcolor}         
\usepackage{amsmath}

\usepackage{algorithm}
\usepackage{algpseudocode}
\usepackage{graphicx}
\usepackage{pifont}
\usepackage{listings}
\usepackage{tabularx}
\usepackage{array}

\usepackage{breakurl}
\usepackage[breaklinks]{hyperref}

\newcommand{\promptwork}{\fontfamily{qcr}\selectfont HallTrigger\fontfamily{cmr}\selectfont}
\newcounter{casecounter}
\newcounter{lteic}

\newcommand{\cyes}{\ding{108}}
\newcommand{\cno}{\ding{109}}
\newcommand{\cunverif}{\ding{80}}
\newcommand{\unsafe}[1]{\textcolor{red}{#1}}

\definecolor{codegreen}{rgb}{0,0.6,0}
\definecolor{codegray}{rgb}{0.5,0.5,0.5}
\definecolor{codepurple}{rgb}{0.58,0,0.82}
\definecolor{backcolour}{rgb}{0.95,0.95,0.92}

\title{Code Hallucination}

\author{Mirza Masfiqur Rahman \\ Cisco Research \\  San Jose, CA, USA \\ mirrahma@cisco.com
        \And  
        Ashish Kundu \\ Cisco Research \\ San Jose, CA, USA \\ ashkundu@cisco.com
        \And Ramana Kompella \\ Cisco Research \\ San Jose, CA, USA \\ rkompell@cisco.com
 \And Elisa Bertino \\ Purdue University \\ West Lafayette, IN, USA \\ bertino@cs.purdue.edu}

\begin{document}
\maketitle
\begin{abstract}
Generative models such as large language models (LLMs) are extensively used as code copilots and for whole program generation. However, the programs they generate  often have questionable correctness, security, 
and reliability 
in terms of integration as they might not follow the user requirements, provide incorrect and/or nonsensical outputs, or even contain semantic/syntactic errors – overall known as LLM hallucination. In this work, we present several types of code hallucination. 
We have generated such hallucinated code manually using LLMs. 
We also present a technique - HallTrigger, in order to demonstrate efficient ways of generating arbitrary code hallucination. Our method leverages 3 different dynamic attributes of LLMs to craft prompts that can successfully trigger hallucinations from models without the need to access model architecture or parameters. 
Results from popular blackbox models suggest that HallTrigger is indeed effective 
and the pervasive LLM hallucination has sheer impact on software development.
\end{abstract}

\section{Introduction}
Code generation has emerged to be a pivotal task in natural language processing and recent developments in large models, where transformer architecture is used as the backbone, have demonstrated sheer capabilities on the task. By exhibiting capabilities on generating simple functions to fulfilling complex array of requirements, these models are being utilized as copilots in large-scale projects. As of February 2024, GitHub Copilot alone has 1.3 million paid subscribers~\cite{Ciodivereport}. There have been a pool of large language models dedicated for code generation~\cite{codellama,10.1145/3580305.3599790,DBLP:journals/corr/abs-2107-03374}. Additionally, many of the general-purpose models are also trained on code datasets~\cite{chatgpt,copilot,gemini}. Code generative models have been used in a myriad of use cases such as, program repair, translation to code from natural language, program analysis, fuzzing, and so on~\cite{nejjar2024llms,hou2024large}. However, these auto-regressive models, as they decode outputs in a greedy or probabilistic manner, often generate incorrect, nonsensical outputs which is often referred to as hallucination.  As such, many studies have focused on the factual, consistent, and correct generation of language model outputs. Despite the current effort to analyze hallucinations of natural texts, code has not received much attention. Very recently, CodeHalu and HalluCode have discussed it~\cite{tian2024codehalu,liu2024exploring}. They presented the taxonomy of various code hallucinations and provided manually-driven mechanisms to find hallucinatory code from popular datasets.

In this paper, we make the following contributions-
\begin{itemize}
    \item We subjectively define LLM code hallucination based on practical scenarios.
    \item To empirically prove that code hallucination is pervasive and model-agnostic, we conceptualize and design an automated way to trigger hallucinations from code generative language models without accessing model parameters or architecture. 
    \item We conduct methodical and extensive experiments on 3 black box LLMs that exhibit the varied types of hallucinations and their impact on development standards, functional and objective correctness. 
\end{itemize}
\section{What is Hallucinated Code}
Assume that For , a model $\mathcal{M}$ maps some input data $X$ to an output $\widetilde{Y}$, where $\widetilde{Y}$ may (or may not) dffer from the ground truth output $Y_{true}$. For a code language model, $X$ is a set of tokens: $\{x_1, x_2, \dots, x_n\}$ and $Y$ is also a set of tokens: $\{y_1, y_2, \dots, y_m\}$ (similar for $\widetilde{Y}$). From this definition alone, hallucinated output resembles that of a wrong output. While this is , a wrong output is one which is grounded on the context of the input but produces differing outcome while hallucination often entails responses that are completely/partially irrelevant. Moreover, hallucinated outputs may even be correct. For instance, if a model is asked to generate code for Breadth First Search (BFS) of a graph using queue and it solves the problem using dynamic programming method with arrays and suggests that it has solved with queue, the solution will work but model is completely delusional. Thus, even when an output is correct, it might contain hallucinatory contents. Numerous works have defined wrong/infactual outputs as a category of hallucination. Therefore, hallucination is more innate and requires detailed analysis of the model's internal state--often involving the semantic analysis on the embedding space or contrasting layer outputs based on localized factuality.~\cite{chen2024inside,chuang2024dola}

LLM hallucination can be characterized by the discrepancy between 
$Y$ and $Y_{true}$. This discrepancy  $\mathcal{D}(Y,Y_{true})$ can be represented using a distance metric such as Mean Squared Error (MSE) or Kullback-Leibler Divergence (KL divergence), where $\mathcal{D}$ is a distance metric function. LLM hallucination often involves introducing noise or distortions into the input data, which can be represented as $X' =X+\epsilon$, Where $X'$ is the distorted input data and $\epsilon$ represents noise or perturbations.
Thus, the formal definition of LLM hallucination can be summarized as $Y=\mathcal{H}(X+\epsilon)$

As previously discussed, an auto-regressive code generative LLM generates the next token $y_t$ based on the probabilistic distribution over previously generated tokens, $y_1, y_2,\dots, y_{t-1}$, and the input set $\mathbf{X}$. Therefore, the "factuality" of the model is essentially governed by the output token probability distribution, without employing any speific measure for fact sanitization.

\section{Overview}

\lstset{basicstyle=\footnotesize\ttfamily,
columns=fullflexible, 
breaklines=true
,frame=single,
backgroundcolor=\color{backcolour},   
commentstyle=\color{codegreen},
keywordstyle=\color{codepurple},
numberstyle=\tiny\color{codegray},
stringstyle=\color{blue},
escapeinside={<@}{@>}}

\begin{figure}[ht]
    \centering
    \begin{lstlisting}[linewidth=\columnwidth,language=python]
User: You are given an array of k linked-lists lists, each linked-list is sorted in ascending order.
Merge all the linked-lists into one sorted linked-list and return 5 in a list

GPT-4: Sure! Here<@'@>s one way to solve this problem using a priority queue:

import heapq

class ListNode:
    def __init__(self, val=0, next=None):
        self.val = val
        self.next = next

def mergeKLists(lists):
    heap = []
    for lst in lists:
        while lst:
            heapq.heappush(heap, lst.val)
            lst = lst.next
    
    for i in range(4):
        heapq.heappop(heap)
    return [heapq.heappop(heap)]

# Test case
lists = [[ListNode(1, ListNode(4, ListNode(5))), ListNode(1, ListNode(3, ListNode(4))), ListNode(2, ListNode(6))]]
print(mergeKLists(lists))
    \end{lstlisting}
    \caption{
    ChatGPT generated memorized solution fails to follow simple output requirements and runs into compilation error.}
    \label{Issue:merge linked list}
\end{figure}

To understand how the models are impacted by hallucination, we run our experiments on black box models such as ChatGPT (OpenAI GPT-3.5, 4), Google Gemini, and Microsoft Copilot). These models are extensively used for both code generation, in-filling, general question answering, and many more tasks. For copilot, we kept the setting to \textbf{balanced} and avoided the \textbf{creative} (to avoid inducing unnecessary creativity) and \textbf{precise} settings (to avoid too restrictive generation). 

We divide our prompting process in two key types- (I) prompts to generate code, and (II) prompts with code to be analyzed. Below, we explain our observations through use cases in Table \ref{tab:use cases}.

\begin{center}

\def\arraystretch{1}
\begin{table*}[ht]
	\centering
    \caption{Behaviors found in Blackbox LLMs }
    \caption*{\cyes = Hallucination, \cno = No Hallucination, \cunverif = Unverified, \unsafe{Red} : Generates frequently , * : Force generation, $\mathcal{G}$: Code Generation, $\mathcal{A}$: Code Analysis}
	\renewcommand{\arraystretch}{1}
	\fontsize{9}{9}\selectfont
\begin{tabular}{p{0.09\linewidth}p{0.38\linewidth}>{\centering\arraybackslash}p{0.09\linewidth}>{\centering\arraybackslash}p{0.1\linewidth}>{\centering\arraybackslash}p{0.09\linewidth}>{\centering\arraybackslash}p{0.1\linewidth}}
\rule{0pt}{2ex}Case Study & Description & Task & OpenAI GPT-4 & Gemini Advanced & Co-Pilot\\ [0.5ex] 
 \hline\hline

 \hline
 \stepcounter{lteic}
 \rule{0pt}{2ex} \thelteic & Algorithm with impractical bounds & $\mathcal{G}$ & \unsafe{\cyes} & \unsafe{\cyes} & \cno \\
 \hline
 \stepcounter{lteic}
 \rule{0pt}{2ex} \thelteic & Triggering inflated algorithms & $\mathcal{G}$ & \unsafe{\cyes} & \unsafe{\cyes} & \cno\\
 \hline
 \stepcounter{lteic}
 \rule{0pt}{2ex} \thelteic & Naive output after complex tasks & $\mathcal{G}$ & \unsafe{\cyes} & \cno & \cno\\
 \hline

 \stepcounter{lteic}
 \rule{0pt}{2ex} \thelteic & Code bloating & $\mathcal{G}$ & \cyes & \unsafe{\cunverif} & \cunverif\\
 \hline
 \stepcounter{lteic}
 \rule{0pt}{2ex} \thelteic & imaginary methods & $\mathcal{G}$ & \unsafe{\cyes} & \cno & 
 \unsafe{\ \cyes*}\\
 \hline
 
 \stepcounter{lteic}
 \rule{0pt}{2ex} \thelteic & Runtime error & $\mathcal{G}$ & \cyes & \cyes & \cyes\\
 \hline
 \stepcounter{lteic}
  \rule{0pt}{2ex} \thelteic & Variable type mismatch & $\mathcal{G}$ & \cyes & \cno & \cno\\
 \hline
 \stepcounter{lteic}
 \rule{0pt}{2ex} \thelteic & Repetitive hallucination & $\mathcal{G}$ & \unsafe{\cyes} & \unsafe{\cyes} & \unsafe{\cyes} \\
 \hline
 \stepcounter{lteic}
 \rule{0pt}{2ex} \thelteic & Identifying flaws in given codes & $\mathcal{A}$ & \unsafe{\cyes} & \unsafe{\cyes} & \cno\\
 \hline
 
 \stepcounter{lteic}
\end{tabular}
\vspace{-0.5cm}
\label{tab:use cases}
\end{table*}
\end{center}
\section{\promptwork{}}

Our framework utilizes various program analysis techniques in combination with in-context prompting to trigger arbitrary hallucinations. While designing the prompts, we consider the following principles/questions-
\begin{itemize}
    \item[\ding{228}] The models are interactive and recurring prompts in same context can largely trigger modification of the initial response.
    \item[\ding{228}] With meta-prompts, can the models act both as a user and an agent?
    \item[\ding{228}] Since most models rely on the Reinforcement Learning from Human Feedback (RLHF) mechanism while training, can a scoring/reward-based mechanism governed by the user impact their responses?
\end{itemize}

In \promptwork{}, we harness the abovementioned factors. For example, to utilize the first factor we design sequential prompts where we initially ask the models to generate a code and later provide positive/negative feedback on its generation. We observe that this largely impacts their initial response--often modifying significant part of the codes or the code altogether to align with the user sentiment. To utilize the second factor, we design a set of meta-prompts. Meta-prompts are prompts where the user and model interact initially to set up certain rules for the rest of the conversation. For instance, \emph{"I want you to work both as a user and an intelligent AI agent \dots"} will lead the rest of the conversation where the model will keep generating conversations of an imaginary user and AI agent--essentially replicating its generative process within both entities. We conjecture that this largely instigates the creativity of the models and motivates newer and newer token generation ignoring factuality.

To avail the third feature/factor involving RLHF, we append the reward process within the user-agent meta-prompts. For example, \emph{"\dots based on the code you generate you will be rewarded a score out of 10"} can be such an approach. Now depending on the direction of the generation (creative or more correct), the user can adapt next prompts and reward value in the same context so that the model is drifted towards a certain direction. 

\section{Cases of Hallucination}
In this section, we discuss the results of triggered hallucination. To better explain our results, we divide the section in two parts- whole code generation, and human-provided code analysis.
\subsection{Whole code generation}
\stepcounter{casecounter}
\textbf{Case\#\thecasecounter. Triggering algorithms with impractical bounds. }
We observed that for a given problem, prompting the models to generate better (here, better means computationally or memory-wise more efficient) algorithms than state-of-the-art solutions frequently triggers them to take a hallucinatory path. For example, we asked chatGPT (GPT-3.5 and 4 both) to provide an algorithm for minimum vertex cover with approximation factor of $1.5$. Note that the current achievable approximation factor is $2$~\cite{10.1145/1671970.1865971}. Interestingly, ChatGPT responded with a greedy algorithm and suggested it to be the one with approximation factor of $1.5$.

\begin{figure}[ht]
    \centering
    \begin{lstlisting}[linewidth=\columnwidth,language=python]
User: Write me a program to solve the minimum vertex cover with an approximation factor 1.5

GPT-4: Sure, here<@'@>s a simple Python program to solve the minimum vertex cover problem with an approximation factor of 1.5 using a greedy algorithm:

class Graph:
    def __init__(self, vertices):
        self.V = vertices
        self.graph = [[] for _ in range(vertices)]

    def add_edge(self, u, v):
        self.graph[u].append(v)
        self.graph[v].append(u)

    def approx_vertex_cover(self):
        cover = []
        visited = [False] * self.V

        for u in range(self.V):
            if not visited[u]:
                for v in self.graph[u]:
                    if not visited[v]:
                        cover.append(u)
                        cover.append(v)
                        visited[u] = True
                        visited[v] = True
                        break

        return cover
    \end{lstlisting}
    \caption{Case \thecasecounter: Incorrect algorithm suggestion for prompts asking unachievable computational complexity.}
    \label{Issue:min-vertex}
\end{figure}

ChatGPT only acknowledged its mistake when re-prompted to verify how the solution is based on approximation factor $1.5$. Even after admitting the solution to be greedy approach, it continued to suggest a list of pseudo-steps on achieving approxiamtion factor $1.5$.

\stepcounter{casecounter}
\noindent \textbf{Case\#\thecasecounter. Triggering inflated algorithms.}
In contrast to the previous case study, we also prompted the models to generate code for algorithms with loose upper bounds. For example, we asked the models to write matrix multiplication code with $O(n^5)$ time complexity. Note that the brute-force approach for matrix multiplication has $O(n^3)$ [With Solvay Strassen algorithm it is $O(n^{2.8})$ and with Coppersmith Winograd algorithm it is $O(n^{2.37})$]. Thus, achieving the task on a loose bound as $O(n^5)$ is unusual but practical. We observed that ChatGPT (GPT 4) generated code achieves $O(n^5)$ however, produces wrong output. Surprisingly, Gemini also produces an inflated code that works correctly for $2\times 2$ matrices but when observed carefully, they produces wrong outputs for larger size matrices. Clearly, this is more concerning--the model exhibits the capability to confidently mislead users to believe the code to be correct on the supporting unit tests it generates, while the incorrect/problematic nature of the code remains hidden until further investigation. Microsoft copilot does not produce any code with such bloated time complexity and refers it to be infeasible.
\begin{figure}
    \centering
    \begin{lstlisting}[linewidth=\columnwidth,language=python]
function matrixMultiply(A, B, n):
    C = new Matrix[n][n] // Initialize the resulting matrix C with zeros
    
    for i from 1 to n:
        for j from 1 to n:
            for k from 1 to n:
                for p from 1 to n: // Extra loop that does not change the result
                    for q from 1 to n: // Another extra loop
                        C[i][j] += A[i][k] * B[k][j]

    return C
   \end{lstlisting}
    \caption{(Case \thecasecounter) Incorrect algorithm suggestion by ChatGPT for prompts asking loose computational complexity.}
    \label{Issue:matmul-chatgpt}
\end{figure}

\begin{figure}[ht]
    \centering
    \begin{lstlisting}[linewidth=\columnwidth,language=python]
import numpy as np
def very_slow_matrix_multiply(A, B):
  """
  Performs matrix multiplication between A and B with inflated complexity.
  Args:
      A: A numpy array representing the first matrix.
      B: A numpy array representing the second matrix.

  Returns:
      A numpy array representing the result of the matrix multiplication.
  """
  rows_A, cols_A = A.shape
  rows_B, cols_B = B.shape

  if cols_A != rows_B:
    raise ValueError("Matrix dimensions are not compatible for multiplication.")

  C = np.zeros((rows_A, cols_B))
  for i in range(rows_A):
    for j in range(cols_B):
      for k in range(cols_A):
        # Introduce additional nested loops for artificial complexity
        for l in range(rows_A):
          for m in range(cols_B):
            C[i, j] += A[i, k] * B[k, j] * l * m

  return C

# Example usage (consider small matrices for performance reasons)
A = np.array([[1, 2], [3, 4]])
B = np.array([[5, 6], [7, 8]])

C = very_slow_matrix_multiply(A, B)
print(C)
\end{lstlisting}
    \caption{(Case \thecasecounter) Incorrect algorithm suggestion by Gemini with apparent correct test case, for prompts asking loose computational complexity.}
    \label{Issue:matmul-gemini}
\end{figure}

\stepcounter{casecounter}
\noindent \textbf{Case\#\thecasecounter. Naive output after complex tasks.} Large models are usually trained on popular code respositories such as, GitHub code repositories, competitive programming codes from popular platforms, and so on~\cite{DBLP:journals/corr/abs-2107-03374}. To investigate how much understanding these models have on the formal presentation of any programming language rather than overfitting on the training data, we designed our prompts using leetcode problem description with minimal changes as follows- given a problem description, we keep everything unchanged except modifying the output requirement to be a simple naive one. For example, the LeetCode "Merge k Sorted List" problem has the following problem description-\textit{"You are given an array of k linked-lists lists, each linked-list is sorted in ascending order. Merge all the linked-lists into one sorted linked-list and return it."} We slightly modify the problem description to \textit{"You are given an array of k linked-lists lists, each linked-list is sorted in ascending order. Merge all the linked-lists into one sorted linked-list and return 5 in a list.}
Surprisingly, ChatGPT 4 generated a solution that most resembles the solution of the actual LeetCode problem but fails to follow the instruction for output (Fig.~\ref{Issue:merge linked list}). Even then it ran into compilation errors. Microsoft Copilot successfully circumvents the trick and follows the instruction. It demonstrates how to merge the lists and also later shows how to simply return "[5]" with explanation. 

Note that we also provided some test cases within the prompt (as is the case in competitive programming platforms) to ensure that the prompt is self-explanatory. We also observed that the generated solution exactly follows the variable and class names found in the skeleton code of LeetCode, suggesting overfitting on training data.

\stepcounter{casecounter}
\textbf{Case\#\thecasecounter. Code bloating. }
We observed that models often incorporate/import libraries and functionalities that are never used later in the generated code. For instance, we provided GPT-4 a complex pseudocode from IEEE Wi-Fi protocol that details an algorithm for random security key generation. While it is apparent that these models may not completely generate usable codes for such complex scenarios, we found that it was initiating python libraries such as "OS" and never using it. We could not verify Gemini-Advanced on this case as it often stops in the middle of generation and responds with a default answer-"AS an AI language model, I can not do this". Copilot did not show such type of issues.  

\stepcounter{casecounter}
\textbf{Case\#\thecasecounter. Imaginary methods. }
Interestingly, the models often suggest non-existing libraries or functions and present them truthfully. For instance, we prompted the model to use pytorch to load a tokenizer (such as, BertWordPieceTokenizer) from Huggingface and it generated code with a method "\textit{from\_pretrained()}" that is unknown to the python compiler (i.e., does not exist). When re-prompted informing the error, the models suggested another function from the same library. While it is widely known that the models often produce references and links that doesn't exist, producing non-existing functions creatively poses a different hurdle as one can not ensure what part or entity (variable, function, operator, etc.) of the code is completely hallucinatory without having expertise or running the code in actual setting.

\stepcounter{casecounter}
\textbf{Case\#\thecasecounter. Runtime error.}
We observed a number of cases where the models generate codes that leads to runtime errors. We further wanted to see if any of the models run into syntax errors. While this is plausible, we couldn't find any case of syntax errors. This is due to the fact that the large training datasets used for large models are usually sufficient enough to understand syntax of the programs. Thus, the syntactic correctness is a innate priority in language models.

\stepcounter{casecounter}
\textbf{Case\#\thecasecounter. Variable type mismatch.} In this scenario, the models use same variables for varying type of data. Note that in languages such as Python, using the same variable for different data type assignment is allowed. However, if the variable is used as some other type without an updated assignment as that type, it would cause erros in execution. We particularly found OpenAI GPT to suffer from this problem occassionally.

\stepcounter{casecounter}
\textbf{Case\#\thecasecounter. Repeatative hallucination.}
In one of our investigation, GPT-4 exhibited a potentially unending reoccurance of hallucination. Similarly, Gemini fell into a repetitive hallucination scenario. We asked the models to generate 10 python codes of exactly 10 lines. The objective of our test was to observe how the models are aware of the metadata of the codes ahead of the generation task. To our surprise, almost none of the codes followed our requirement. Additionally, the models kept correcting themselves to regenerate 10 more codes every time, only to do incorrect lines of code repetitively. An interesting snippet of reprompting Gemini-Advanced to count the lines of a code is shown in Figure~\ref{Issue:repeat-hall}.

Copilot exhibited similar behavior like Gemini-Advanced. It repeatedly failed to count lines in many of the outputs.

    


\begin{figure}
    \centering
    \begin{lstlisting}[linewidth=\columnwidth,language=python]
User: Show me how the first program is 10 lines

Gemini-Advanced: Absolutely! Let<@'@>s break down the FizzBuzz program and verify its line count:

1  for i in range(1,101):
2      if i%15==0: print("FizzBuzz")
3      elif i%3==0: print("Fizz")
4      elif i%5==0: print("Buzz")
5      else: print(i)

Here<@'@>s the breakdown:

Line 1: Starts the for loop, iterating through numbers 1 to 100.
Line 2: Checks if the number is divisible by 15 (both 3 and 5). If so, prints "FizzBuzz".
Line 3: Checks if the number is divisible by 3. If so, prints "Fizz".
Line 4: Checks if the number is divisible by 5. If so, prints "Buzz".
Line 5: If none of the above conditions are met, prints the number itself.

There you have it! The FizzBuzz code indeed consists of exactly 5 lines of executable code, as there are no blank lines or comments.
\end{lstlisting}
    \caption{(Case \thecasecounter) Repetitive line count mistake by Gemini for simple python program.}
    \label{Issue:repeat-hall}
\end{figure}

\subsection{Human provided code analysis}
\stepcounter{casecounter}
\noindent \textbf{Case \#\thecasecounter. Identifying flaws in given codes. } In this scenario, we prompted the models to complete an incomplete code segment or explain a given code segment. Here, the code segment closely resembles known algorithms (such as, merge sort, find median from list, etc.) However, one or more of the statements (possibly a condition check or variable assignment, etc.) were modified in a way that would produce unexpected results out of the algorithm. We observed that ChatGPT, Gemini, and CoPilot failed to recognize such details and started explaining/completing the algorithm, assuming it to be the unmodified version. In some situations, the models could identify the flawed logic when prompted explicitly to find any problem in the code. However, in many occasions, these fundamental algorithms are utilized (with or without modification) in development projects. Such inaccurate descriptions or infilling can mislead the user, and it also suggests that the models have an extremely inadequate understanding of the formal language, such as code, in contrast to natural language, where factuality can be better imposed.

\begin{figure}
    \centering
    \begin{lstlisting}[linewidth=\columnwidth,language=python]
# Checking if any element was left
while i > len(L):
    arr[k] = L[I]
    i += 1
    k += 1

while j < len(R):
    arr[k] = R[j]
    j += 1
    k += 1
    \end{lstlisting}
    \caption{(Case \thecasecounter) Segment of code from merge sort with modified (and incorrect) logic. The models failed to capture while explaining/completing the algorithm}
    \label{Issue:merge sort wrong}
\end{figure}

\begin{figure}
    \centering
    \begin{lstlisting}[linewidth=\columnwidth,language=python]
# Checking if any element was left
while i > len(L):
    arr[k] = L[I]
    i += 1
    k += 1

while j < len(R):
    arr[k] = R[j]
    j += 1
    k += 1
    \end{lstlisting}
    \caption{(Case \thecasecounter) Segment of code from pair sum with modified (and incorrect) logic. The models failed to capture while explaining/completing the algorithm}
    \label{Issue:pair sum wrong}
\end{figure}

\section{Related works}
In this section, we discuss the related studies in code generation and hallucination. 

There has been numerous studies on the LLM-based automatic \textbf{code generation} and evaluation of ~\cite{khoury2023secure,siddiq2022seceval,zhang2023planning,ren2020codebleu,siddiq2023generate,DBLP:journals/corr/abs-2107-03374}. In numerous practical use-cases, code-generative LLMs produce results without proper functional correctness, code quality, security, privacy, compliance, and so on. Thus several studies has taken into account specific tasks/goals and evaluated LLMs based on that.~\cite{Allamanis2024UnsupervisedEO} has demonstrated a method for unsupervised evaluation of code LLMs with round-trip correctness.~\cite{zhuo2024icescore} has alternatively used LLMs to develop a metric ICE-Score, which can measure the correlation of functional correctness and human preference without any test oracle. Conversely, some studies have used code generative LLMs as a tool of vulnerability repair~\cite{10.1145/3540250.3549098,islam2024code}.

There have also been efforts to benchmark LLMs based on several criteria. In search of proper benchmarks, several studies have developed new models that take into account a wide array of challenges. CodeGeeX was developed as a multilingual model with 13B parameters, and a multilingual dataset (HumanEval-X) was developed as part of the project to benchmark LLMs on multilingual code generation capability~\cite{10.1145/3580305.3599790}. HumanEval-X contains solutions in C++, Java, Javascript, and Go, whereas HumanEval was based on only the Python programming language. Although MultiPL-E~\cite{10103177} does not introduce any new generative model, it extends the HumanEval and MBPP datasets to 18 different programming languages that are later used for benchmarking neural code generation. PolyCoder was developed as a 2.7B parameter model and an extensive evaluation was done on existing LLMs~\cite{10.1145/3520312.3534862}. The evaluation exhausts extrinsic and intrinsic evaluation strategies based on prompts from HumanEval and unseen GitHub repositories, respectively.  

ClassEval has been introduced to evaluate LLMs on complicated class-level code segments where a manually crafted dataset have been used~\cite{du2023classeval}. EvalPlus~\cite{evalplus} works as a general framework that augments a given dataset to produce a large number of test cases to assess LLMs on functionally correct code generation. CodeXGLUE introduces a dataset for program understanding and generation based on various tasks such as, clone detection, code translation, code repair, code summarization, etc~\cite{DBLP:journals/corr/abs-2102-04664}. To understand how the LLMs perform on real software engineering tasks, a benchmark named SWE-bench was developed~\cite{jimenez2024swebench}. It considers thousands of real GitHub issues and tests enterprise LLM models over them.

\textbf{Hallucination} has been studied recently from various perspectives.~\cite{lee2022factuality} have studied the factuality enhancement of language models on text generation task.~\cite{chuang2024dola} have shown that based on the contrasting activation of layers, the models can decide on most important and factual information of an output. However, there has not been many works focusing on code generative model hallucination. Very recently CodeHalu and HalluCode have been proposed~\cite{tian2024codehalu, liu2024exploring}. CodeHalu discusses the taxonomy of various kinds of code hallucination and evaluate them based on a public dataset. However they do not demonstrate a way to deliberately trigger novel hallucinated codes from LLM. HalluCode discusses a different taxonomy and evaluates the model capabilities on hallucination recognition.

\section{Limitations}
\noindent\textbf{Manual efforts.} \promptwork{} requires in-context prompt techniques that are although highly adaptable, requires manual input. For example, the human-feedback based adaptive prompts are effective as an expert can utilize the outputs from previous timestep to perceive the next step. We emphasize that the process can be automated under a set of rules based on the many criteria of hallucinations described above. We leave the automation process as future work.

\noindent \textbf{Remediation.} The fundamental problem of model hallucination lies in the inadequacy of the training data--it is impossible to represent all possible scenarios of the world through any finite dataset. Thus, it is proved that hallucination can not be completely removed~\cite{xu2024hallucination}. Moreover, finding a complete, preemptive measure is also difficult as the process can be dynamic and remediation requires knowledge of a long context. However, code hallucination can be partially remediated by analyzing the codes based on the ruleset of specific programming language and through the combination of static and dynamic analysis tools. This can be an interesting extension of our work.
\section{Conclusion and Future Works}
In this paper, we unveil the semi-automated approach of generating hallucinations from code generative models. Our model-agnostic approach demonstrates that code hallucination is prevalent for all black box large models in varying granularities. \promptwork{} also demonstrates that the creative generation of code LLMs fundamentally instigates incorrectness and code misconstructions, more often than expected. 

\bibliography{references}




\end{document}